\documentclass[11pt,a4paper]{article}
\usepackage[hyperref]{acl2020}
\usepackage{times}
\usepackage{latexsym}
\usepackage{tcolorbox}
\usepackage[utf8]{vietnam}
\usepackage{graphicx}
\usepackage{microtype}

\usepackage{microtype}

\aclfinalcopy 

\title{ReINTEL Challenge 2020: Exploiting Transfer Learning Models \\for Reliable Intelligence Identification on Vietnamese Social Network Sites}

\author{Kim Thi-Thanh Nguyen$^{1,2}$, Kiet Van Nguyen$^{1,2}$\\
  $^{1}$University of Information Technology, Ho Chi Minh City, Vietnam\\
  $^{2}$Vietnam National University, Ho Chi Minh City, Vietnam \\
  \texttt{18520963@gm.uit.edu.vn}, \texttt{kietnv@uit.edu.vn} \\}
\date{}

\begin{document}
\maketitle
\begin{abstract}
This paper presents the system that we propose for the Reliable Intelligence Identification on Vietnamese Social Network Sites (ReINTEL) task of the Vietnamese Language and Speech Processing 2020 (VLSP 2020) Shared Task. In this task, the VLSP 2020 provides a dataset with approximately 6,000 training news/posts annotated with reliable or unreliable labels, and a test set consists of 2,000 examples without labels. In this paper, we conduct experiments on different transfer learning models, which are bert4news and PhoBERT fine-tuned to predict whether the news is reliable or not. In our experiments, we achieve the AUC score of 94.52\% on the private test set from ReINTEL’s organizers.

\end{abstract}

\section{Introduction}

With the explosion of The Fourth Industrial Revolution in Vietnam, SNSs such as Facebook, Zalo, Lotus have attracted a huge number of users. SNSs have become an essential means for users to not only connect with friends but also freely share information and news. In the context of the COVID-19 pandemic, as well as prominent political and economic events that are of great interest to many people, some people tend to distribute unreliable information for personal purposes. The discovery of unreliable news has received considerable attention in recent times. Therefore, VLSP opens ReINTEL \cite{le2020reintel} shared-task with the purpose of identifying being shared unreliable information on Vietnamese SNSs. 

Censoring news to see if it is trustworthy is tedious and frustrating. It is sometimes difficult to determine whether the news is credible or not. Fake news discovery has been studied more and more by academic researchers as well as social networking companies such as Facebook and Twitter. Many shared-task to detect rumors were held, such as SemEval-2017 Task 8: Determining rumour veracity and support for rumours  \cite{derczynski-etal-2017-semeval} and    SemEval-2019 Task 7: RumourEval, Determining Rumour Veracity and Support for Rumours  \cite{gorrell-etal-2019-semeval}. 

In this task, we focus on finding a solution to categorize unreliable news collected in Vietnamese, which is a low-resource language for natural language preprocessing. Specifically, we implement deep learning and transfer learning methods to classify SNSs news/posts. The problem is stated as:
\begin{itemize}
\item{\textbf{Input}}: Given a Vietnamese news/post on SNSs with the text of news/post (always available), some relative information, and image (may be missing).
\item{\textbf{Output}}: One of two labels (unreliable or reliable) that are predicted by our system.
\end{itemize}
Figure \ref{fig:1} shows an example of this task.

The rest of the paper is organized as follows. In Section ~\ref{sec:Related work}, we present the related work. In Section ~\ref{sec:Methodology}, we explain some proposed approaches and its result. In Section ~\ref{sec:Analysis}, we present the experimental analysis. Finally, Section ~\ref{sec:Conclusion and future work} draws conclusions and future work.

\section{Related work}
\label{sec:Related work}

\citet{DBLP:journals/corr/RuchanskySL17} used a hybrid model called CSI to categorize real and fake news. The CSI model includes three components: capture, source, and integrate. The first module is used to detect a user's pattern of activity on news feeds. The second module learns the source characteristics of user behavior. The last module combines both previous modules to categorize news is real or fake. The CSI model does not make assumptions about user behavior or posts, although it uses both user-profiles and article data for classification.

\citet{DBLP:journals/corr/abs-1910-14353} focused on improving the results of the Fake News Challenge Stage 1 (FNC-1) stance detection task using transfer learning. Specifically, this work improved the FNC-1 best performing model adding BERT  \cite{DBLP:journals/corr/abs-1810-04805} sentence embedding of input sequences as a model feature and  fine-tuned XLNet \cite{DBLP:journals/corr/abs-1906-08237} and RoBERTa  \cite{DBLP:journals/corr/abs-1907-11692} transformers on FNC-1 extended dataset.  

\begin{figure}
  \centering
    \begin{tcolorbox}
    \textbf{Id}: 0.\\
    \textbf{User id}: 2167074723833130000. \\
    \textbf{Post message}: Cần các bậc phụ huynh xã Ngũ Thái lên tiếng, không ngờ xã mình cũng nhận thịt nhiễm sán...
Cho các cháu Mầm non ăn uống thế này thật vô nhân tính!
VTV đăng tin rồi nhé các anh chị.\\
English translation:    \emph{Needing the parents of Ngu Thai commune to speak up, astonishing my commune accept contaminated meat ...
Feeding preschool children like this is so inhumane!
VTV posted the news, guys.}\\
    \textbf{Timestamp post}: 1584426000.  \\
    \textbf{Number of post's like}: 45. \\
    \textbf{Number of post's comment}: 15. \\
    \textbf{Number of post's share}: 8.\\
    \textbf{Label}: 1 (unreliable).\\
    \textbf{Image}: NAN.
    \end{tcolorbox}
  \caption{\label{fig:1}An example extracted from the dataset.}
\end{figure}

\section{Approaches}
\label{sec:Methodology}
In this study, we concentrate on SOTA models, including deep neural network models and transfer learning models.
\subsection{Experimental approaches}
\subsubsection{Deep neural network models} 
In studying the fundamental theories and methods of detecting fake news, \citet{10.1145/3395046} have come up with some fundamental theories of detecting fake news. The authors wrote, \emph{"theories have implied that fake news potentially differs from the truth in terms of, e.g., writing style and quality (by Undeutsch hypothesis)"}. Therefore, we choose text-feature as the primary input of our experimental models.
Firstly, we run deep learning models like Text CNN \cite{DBLP:journals/corr/Kim14f}, BiLSTM \cite{DBLP:journals/corr/ZhouQZXBX16} combine with some pre-trained word embedding models such as FastText\footnote{https://fasttext.cc/docs/en/crawl-vectors.html} \cite{DBLP:journals/corr/BojanowskiGJM16} and PhoW2V\footnote{https://github.com/datquocnguyen/PhoW2V} \cite{tuan-nguyen-etal-2020-pilot} to predict the credibility of news. The results of this approach get an AUC score of 0.84 to 0.86, as shown in Table \ref{tab:1}.
We also plan to experiment with incorporating other features that ReINTEL's organizers provide, such as user id, the number of likes, shares, comments, and image, but the lack of information (shown in Table \ref{tab:2}) leads to enormous dynamic causes us to ignore this approach.

\begin{table}
\centering
\begin{tabular}{lr}
\hline \textbf{Model} & \textbf{AUC} \\ \hline
Text CNN + FastText & 0.865996  \\
Text CNN + PhoW2V & 0.846567  \\
BiLSTM + FastText & 0.863183 \\
BiLSTM + PhoW2V & 0.854487\\
\hline
\end{tabular}
\caption{\label{tab:1} Experimental results of deep neural models on public test set.}
\end{table} 

\begin{table}
\scriptsize
\begin{tabular}{lrrr}
\hline \textbf{Feature name}& \textbf{Train set}&\textbf{Public test set}& \textbf{Private test set}\\
\hline
Id&0&0&0\\
User name&0&0&0\\
Post message&1&0&0\\
Timestamp post&96&28&34\\
Number of like&115&41&616\\
Number of comment&10&7&677\\
Number of share&725&280&742\\
Label&0&0&0\\
Image&3,085&1,148&1,138\\
\hline
\end{tabular}
\caption{\label{tab:2}Statistics of missing values in the dataset.}
\end{table}

\subsubsection{BERT and RoBERTa for Vietnamese}
One of the problems of deep learning is its massive data requirements as well as the need for computing resources. This has spurred the development of large models and transfer learning methods. \citet{nguyen2020finetuning} presents two BERT fine-tuning methods for the sentiment analysis task on datasets of Vietnamese reviews and gets slightly outperforms other models using GloVe and FastText. \citet{DBLP:journals/corr/abs-1901-11504} fine-tuned  BERT under the multi-task learning framework and obtains new state-of-the-art results on ten NLU tasks, including SNLI, SciTail, and eight out of nine GLUE tasks, pushing the GLUE benchmark to 82.7\% (an improvement of 2.2\%)\footnote{As of February 25, 2019 on the latest GLUE test set}. Therefore, we attempt to fine-tune PhoBERT\footnote{https://github.com/VinAIResearch/PhoBERT} \cite{nguyen-tuan-nguyen-2020-phobert}  and bert4news\footnote{https://github.com/bino282/bert4news}, pre-trained models for Vietnamese which is based on BERT architecture. And transfer learning shows strength in these experiments, we get an AUC score of between 0.92 to almost 0.95, as shown in Table \ref{tab:3}.
\begin{table}
\centering
\begin{tabular}{lrl}
\hline \textbf{Model} & \textbf{AUC} \\ \hline
PhoBERT & 0.932424\\
bert4news & 0.935163 \\
PhoBERT+bert4news & 0.945169 \\
\hline
\end{tabular}
\caption{\label{tab:3} Results of PhoBERT, bert4news, and results that combine these two models on the private test set.}
\end{table}

\subsection{Fine-tuning BERT and RoBERTa for Vietnamese}
After many experiments, we find that the deep learning models do not achieve higher performance than fine-tuned bert4news and PhoBERT. Therefore, we decided to focus on only improving the results on transfer learning methods. Besides, we also try to combine the results of these two models.

The fine-tuning idea is taken from the study \cite{DBLP:journals/corr/abs-1905-05583}. The BERT base model creates an architecture of 12 sub-layers in the encoder, 12 heads in multi-head attention on each sub-layer. BERT input is a sequence of not more than 512 tokens; the output is a set of self-attention vectors equal to the input length. Each vector is 768 in size. The BERT input string represents both single text and text pairs explicitly, where a special token [CLS] is used for string sorting tasks, and a special token [SEP] marks the end position of the single text or the position that separates the text pair. For fine-tuning the BERT architecture for text classification, we concatenated the last four hidden representations of the [CLS] token, which will be passed into a small MLP network containing the full connection layers to transform into the distribution of discrete label values. 

Our fine-tuning process consists of two main steps: tokenize the text content and retrain the model on the dataset. For PhoBERT, we use VNcoreNLP \cite{vu-etal-2018-vncorenlp} library to tokenize content, while for bert4news, we use BertTokenizer.

\section{Experiments}
\label{sec:Analysis}
\subsection{Experimental settings}
In this paper, we conduct various experiments on Google Colab (CPU: Intel(R) Xeon(R) CPU @ 2.20GHz; RAM: 12.75 GB; GPU Tesla P100 or T4 16GB with CUDA 10.1). We fine-tune PhoBERT and bert4news with different parameters as batch size, learning rate, epoch, random seed. To save time and cost, we set batch size 32 for all models. With the same hyperparameter values, distinct random seeds can lead to substantially different results \cite {dodge2020finetuning}. With the above configuration, we spend about 2.40 minutes per epoch for both bert4news and PhoBERT. Table \ref{tab:4} shows the parameter setting and the performance, respectively.
\subsection{Performances over time}
\begin{figure}
\centering
\includegraphics[width=0.5\textwidth]{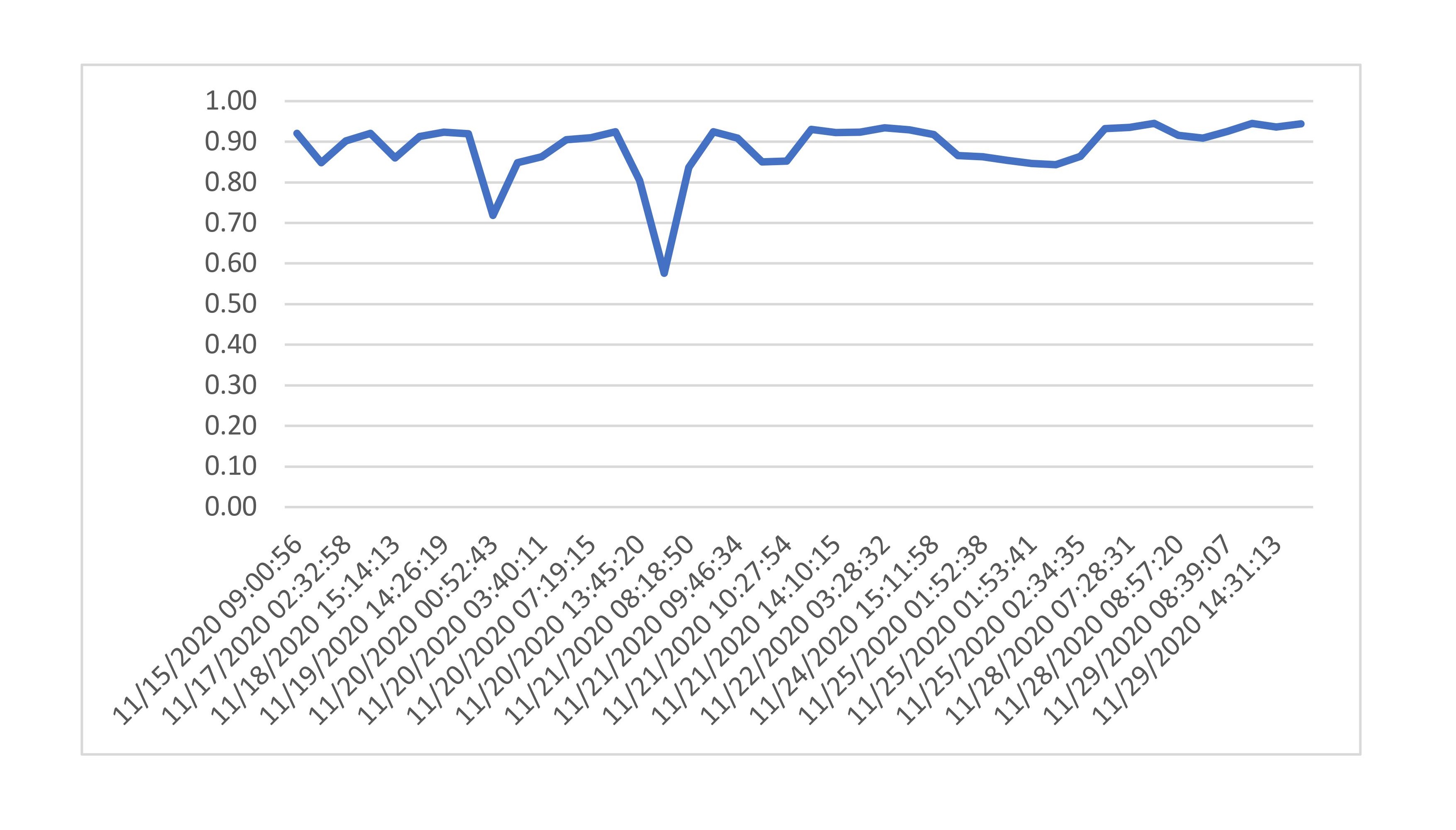}
\caption{\label{fig:2}Performances of the team during the challenging task.}
\end{figure}

Figure \ref{fig:2} shows the results of our testing process. It is easy to see that our results are not stable in the first phase due to trying many methods. Our results are more stable in the later stage of the competition, but there are not many mutations. 
\begin{table}
\scriptsize
\centering
\begin{tabular}{lrrrr}
\hline \textbf{Model} & \textbf{Epochs} & \textbf{Random Seed} & \textbf{Learning Rate} & \textbf{AUC}\\ \hline
PhoBERT  & 5 & 42 & 1.00e-5 & 0.901628\\
PhoBERT  & 6 & 42 & 3.00e-5 & 0.920835 \\
PhoBERT  & 7 & 38 & 2.00e-5 & 0.924961\\
PhoBERT  & 7 & 42 & 2.00e-5 & 0.932424\\
bert4news  & 5 & 42 & 3.00e-5 & 0.930596\\
bert4news  & 6 & 24 & 2.00e-5 & 0.922787\\
\hline
\end{tabular}
\caption{\label{tab:4} Parameter changes lead to a change of results on the public test set.}
\end{table}

\section{Conclusion and future work}
\label{sec:Conclusion and future work}
In summary, we have proposed the following methods for classifying untrustworthy news: combining deep learning model with pre-trained word embedding, fine-tune bert4news, and PhoBERT, combining text, numeric, and visual features. Accordingly, the best result belongs to the transfer learning models when achieving an AUC score of 94.52\% for the combined model of bert4news and PhoBERT. 

In the future, we plan to combine other features offered by ReINTEL’s organizers with transfer learning models due to classifying based on news content alone is not enough \cite{10.1145/3289600.3290994}. While we are doing well in transfer learning, we also aim to build a system for the fast and accurate detection of fake news at the early stages of propagation, which is much more complicated than detecting long-circulated news. Besides, we hope to develop a system to score users based on the news they post and share to reduce unreliable news on Vietnam SNSs.

\bibliography{acl2020}
\bibliographystyle{acl_natbib}
\end{document}